\documentclass[10pt,twocolumn,table,letterpaper]{article}

\usepackage{iccv}              

\definecolor{iccvblue}{rgb}{0.21,0.49,0.74}
\usepackage[pagebackref,breaklinks,colorlinks,allcolors=iccvblue]{hyperref}
\usepackage{xcolor}
\usepackage{float}
\usepackage{amssymb}
\usepackage{multicol}
\usepackage{balance}
\usepackage{lipsum}
\usepackage{graphicx}
\usepackage{amsmath}
\usepackage{booktabs}
\usepackage{subcaption}
\usepackage{multirow} 
\usepackage{makecell} 
\usepackage{array}
\usepackage{adjustbox} 

\def\paperID{8660} 
\def\confName{ICCV}
\def\confYear{2025}

\title{Uncertainty-Driven Expert Control: \\ Enhancing the Reliability of Medical Vision-Language Models}

\author{
    Xiao Liang\textsuperscript{1}, 
    Di Wang\textsuperscript{1,*}, 
    Zhicheng Jiao\textsuperscript{2}, 
    Ronghan Li\textsuperscript{1}, 
    Pengfei Yang\textsuperscript{1}, 
    Quan Wang\textsuperscript{1}, 
    Tat-Seng Chua\textsuperscript{3}
    \\[\medskipamount]
    \textsuperscript{1}The Key Laboratory of Smart Human-Computer Interaction and \\ Wearable Technology of Shaanxi Province, Xidian University, China \\
    \textsuperscript{2}Warren Alpert Medical School, Brown University, USA \\
    \textsuperscript{3}National University of Singapore, Singapore \\
    \texttt{\url{https://github.com/ecoxial2007/Expert-CFG}}
}

\begin{document}
\twocolumn[{
\maketitle

\vspace{-1cm}
\begin{center}
\centering
    \includegraphics[width=\textwidth]{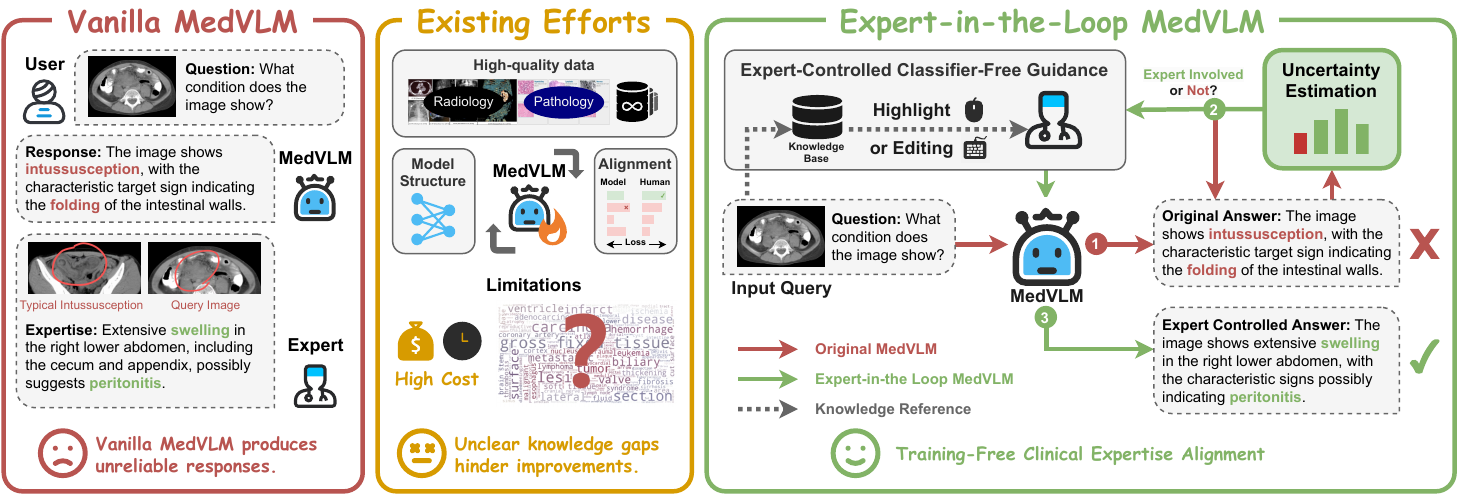}
    \captionof{figure}{\textcolor[rgb]{0.722, 0.329, 0.314}{\textbf{Left}}: Vanilla MedVLM produces unreliable responses due to lack of clinical expertise. \textcolor[rgb]{0.843, 0.608, 0.0}{\textbf{Middle}}: Existing methods mostly rely on training and are costly while still lacking sufficient clinical alignment. \textcolor[rgb]{0.510, 0.702, 0.4}{\textbf{Right}}: Our proposed \textbf{Expert-CFG} utilizes plug-and-play selective expert annotations and classifier-free guidance for effective expertise alignment.}
    \label{fig:motivation}
\label{fig:teaser}
\end{center}
}]

\begingroup\renewcommand\thefootnote{*}
\footnotetext{Corresponding authors.}
\endgroup

\begin{abstract}

The rapid advancements in Vision Language Models (VLMs) have prompted the development of multi-modal medical assistant systems. Despite this progress, current models still have inherent probabilistic uncertainties, often producing erroneous or unverified responses—an issue with serious implications in medical applications. Existing methods aim to enhance the performance of Medical Vision Language Model (MedVLM) by adjusting model structure, fine-tuning with high-quality data, or through preference fine-tuning. However, these training-dependent strategies are costly and still lack sufficient alignment with clinical expertise. To address these issues, we propose an expert-in-the-loop framework named \textbf{Expert}-Controlled \textbf{C}lassifier-\textbf{F}ree \textbf{G}uidance (\textbf{Expert-CFG}) to align MedVLM with clinical expertise without additional training. This framework introduces an uncertainty estimation strategy to identify unreliable outputs. It then retrieves relevant references to assist experts in highlighting key terms and applies classifier-free guidance to refine the token embeddings of MedVLM, ensuring that the adjusted outputs are correct and align with expert highlights. Evaluations across three medical visual question answering benchmarks demonstrate that the proposed \textbf{Expert-CFG}, with 4.2B parameters and limited expert annotations, outperforms state-of-the-art models with 13B parameters. The results demonstrate the feasibility of deploying such a system in resource-limited settings for clinical use.

\end{abstract}    
\section{Introduction}
\label{sec:intro}

Recently, large pre-trained visual language models (VLMs) \cite{llavaLiu2023VisualIT, Ye2023mPLUGOwI2RM, Bai2023QwenVLAV}, such as the latest advancements in GPT-4 \cite{GPT4TROpenAI2023}, have demonstrated impressive performance across various visual and language tasks. However, when applied to the medical domain, these models often underperform due to a lack of domain-specific knowledge \cite{1Yan2023MultimodalCF, 2Jin2024HiddenFB}. Although researchers have developed seemingly effective medical VLMs by instruction fine-tuning with medical multimodal data \cite{Luo2022BioGPTGP, LLaVAMedTALi2023, Moor2023MedFlamingoAM}, these models may generate responses that are factually incorrect or inconsistent with the image content \cite{Liu2024ASO, Huang2023OPERAAH, vcdLeng2023MitigatingOH, lureZhou2023AnalyzingAM} such as that in Figure \ref{fig:motivation}. This severely hampers their applications in the medical field, where high reliability is critical.

To further enhance the performance of MedVLMs and expand their application across different medical domains, researchers have undertaken various efforts that can be broadly divided into two categories: 1) \textbf{Improving MedVLM architectures}, including employing hierarchical image encoding \cite{HighResolution}, utilizing approaches like Mixture-of-Experts \cite{Jiang2024MedMoEMO} or Mamba \cite{Wang2024CXPMRGBenchPA} to better model complex medical knowledge. 2) \textbf{Building high-quality medical visual instruction tuning datasets or datasets aligned with expert preferences}, including HuatuoGPT-Vision \cite{huatuogptv} and BioMed-VITAL \cite{BioMedVITAL}, which aim to improve data quality by incorporating clinician preferences into both data generation and selection processes to ensure better alignment with medical requirements. Although these methods have achieved certain levels of success, several challenges remain. 1) Constructing high-quality medical instruction datasets or conducting additional training often leads to significant costs in terms of time and computational resources. 2) The complexity of clinical expertise makes it challenging to accurately quantify MedVLM's errors. In other words, it is still difficult to determine the specific knowledge deficiencies leading to such unreliability. This complexity hinders the improvements needed for MedVLMs to achieve clinically feasible performance. To address the aforementioned challenges and enable applications in the high-reliability-demanding medical field, we propose an innovative framework called Expert-Controlled Classifier-Free Guidance (\textbf{Expert-CFG}), which is aimed at addressing unreliable responses in existing MedVLMs caused by knowledge deficiencies. Specifically, the construction of this framework requires attention to two research questions:

\begin{itemize}
    \item \textbf{RQ1: How can we determine which MedVLM responses are unreliable?} We utilize predictive entropy to estimate uncertainty in MedVLMs. By identifying responses with high entropy, which is indicative of the model potentially lacking relevant knowledge, we can target these instances for supplementary information. This ensures that expert annotations are applied where they are most needed to correct potential inaccuracies.
    \item \textbf{RQ2: How can we incorporate expert annotations efficiently to enhance output reliability without placing undue burden on experts?} Instead of requiring experts to manually write full annotations, we integrate expert knowledge by retrieving key information and having experts highlight or modify critical content. We then use classifier-free guidance to ensure that MedVLM follows the key expert annotations. This method embeds expert insights into the model with minimal effort, reducing the workload on experts while effectively correcting the model's erroneous responses.

\end{itemize}

Our contributions are summarized as follows:
\begin{itemize}
    \item We propose a training-free framework, \textbf{Expert-CFG}, that efficiently aligns MedVLM with expert annotations, significantly enhancing the reliability of MedVLM without imposing a heavy burden on medical professionals.
    \item We conduct an in-depth analysis of uncertainty and expertise misalignment—factors that remain relatively underexplored in MedVLMs.
    \item The experimental results demonstrate that our framework outperforms existing methods, achieving state-of-the-art performance on three medical visual question answering benchmarks and surpassing the performance of 13B-parameter models using smaller 4.2B models.
\end{itemize}
\section{Related Work}

\textbf{Medical Vision Language Models}. Recent advances in general vision language models, such as Flamingo \cite{Alayrac2022FlamingoAV}, InstructBLIP \cite{Dai2023InstructBLIPTG}, and LLaVA \cite{llavaLiu2023VisualIT}, have significantly improved the integration of vision and language capabilities, enabling more sophisticated image understanding and response generation. Building on these developments, specialized Medical Vision Language Models (MedVLMs), such as LLaVA-Med \cite{LLaVAMedTALi2023}, BioMedGPT \cite{BiomedGPTZhang2023AGV}, Med-Flamingo \cite{Moor2023MedFlamingoAM}, and PMC-VQA \cite{PMCVQAVIZhang2023}, incorporate biomedical visual instruction tuning to meet the unique demands of medical imaging, achieving notable advancements in medical diagnostics and patient care protocols. These advances have led to a diverse array of specialized MedVLMs tailored to specific medical imaging fields, including radiology \cite{RadFM, Bai2024M3DA3}, pathology \cite{Dai2024PALLaVAAL, Seyfioglu2023QuiltLLaVAVI, Chen2024CosteffectiveIL}, and ophthalmology \cite{Gao2023OphGLMTA}. However, due to the inherent complexity of medical data and the probabilistic uncertainty of autoregressive models, MedVLMs inevitably produce factual inaccuracies or responses that are inconsistent with expert diagnoses, which severely impacts their clinical reliability. \\
\textbf{Reliability Assessment of MedVLMs.} Evaluating the reliability of MedVLM outputs is essential given their potential impact on clinical outcomes. In the general domain, various techniques have been developed to assess the factuality of VLMs, including binary object presence questioning \cite{Li2023EvaluatingOH}, automated object selection using language models \cite{Hu2023CIEMCI}, and VQA-based negative statements \cite{Lovenia2023NegativeOP}. Recognizing the critical need for reliability in the medical field, recent benchmarks \cite{Xia2024CARESAC, Royer2024MultiMedEvalAB} have been established to evaluate the performance of MedVLMs. Despite these efforts, concrete guidance on addressing factual inconsistencies is still lacking, and targeted improvements for MedVLMs remain an active area of exploration. \\
\textbf{Improving Accuracy for MedVLMs.} To further enhance the accuracy of MedVLMs and expand their applications across various medical fields, researchers have made diverse efforts: 1) \textbf{Data}: For example, HuatuoGPT-Vision \cite{huatuogptv} uses GPT-4V to reformat general medical data into instruction data that more closely aligns with clinical scenarios, while BioMed-VITAL \cite{BioMedVITAL} incorporates clinician preferences to improve data generation and selection processes; 2) \textbf{Model architecture}: \cite{HighResolution} employs hierarchical image encoding, \cite{He2024PeFoMedPE} uses LoRA, \cite{Jiang2024MedMoEMO} adopts a mixture-of-experts model, and \cite{Wang2024CXPMRGBenchPA} implements Mamba, enabling better adaptability to complex medical knowledge across specialties; and 3) \textbf{Retrieval-augmented approaches}, which have shown significant promise in improving the reliability of tasks such as medical visual question answering \cite{Yuan2023RAMMRB} and report generation \cite{DBLP:conf/mm/LiangZWZLW24}. In particular, RULE \cite{Xia2024RULERM} controls factual risk by calibrating the quantity of retrieved content and fine-tuning selected datasets, while MMed-RAG \cite{Xia2024MMedRAGVM} enhances cross-domain alignment through domain-aware retrieval, adaptive context selection, and preference-based fine-tuning strategies. While these methods have improved the factual accuracy of MedVLMs to some extent, the costs of data construction and fine-tuning remain high, and they do not specifically address issues like detailed factual inconsistencies, which continue to affect overall reliability.

\begin{figure*}[htp]
\centering
\includegraphics[width=1\textwidth]{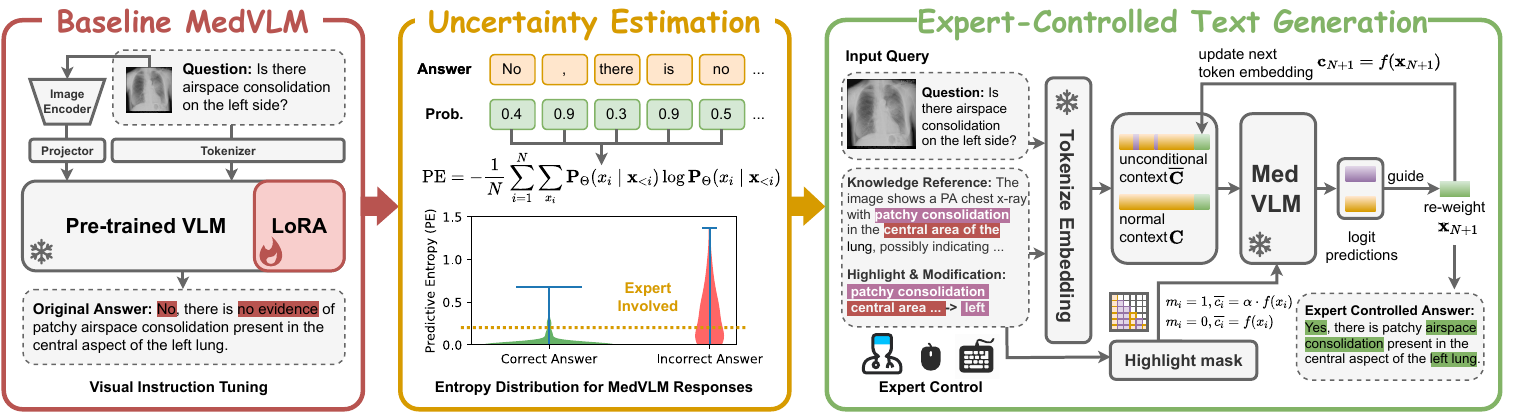}
\caption{\textbf{Overview of our proposed \textbf{Expert-CFG}}, including: (1) Baseline MedVLM with visual instruction tuning; (2) Uncertainty estimation through normalized predictive entropy calculation for MedVLM responses; (3) \textbf{Expert-Controlled Text Generation}, incorporating expert control to adjust embeddings and re-weight predictions.}
\label{fig:overview}
\end{figure*}

\section{Methodology}

Figure \ref{fig:overview} illustrates an overview of the proposed Expert-CFG. Building on a pre-trained MedVLM, the framework operates in three steps: First, it generates an initial answer based on the input image and question. Second, it assesses the answer’s reliability using uncertainty estimation. Third, for low-reliability predictions, it applies expert-controlled text generation, incorporating expert annotations, edits, or knowledge-based retrieval to refine and improve the MedVLM’s output.

\subsection{Baseline Model Construction}
Reliable medical vision language models rely on a strong baseline model. To achieve this, we utilize the powerful Phi-3/Phi-3.5-Vision models \cite{Abdin2024Phi3TR}, which contain 4.2 billion parameters, to construct our MedVLM. We employ 1.3 million meticulously constructed medical visual question answering (VQA) samples \cite{huatuogptv} and follow a two-stage supervised fine-tuning process to refine the model. First, in the stage of \textbf{Alignment VQA}, simple predefined questions are used to mandate that the model comprehensively details the contents of medical images. During this stage, alignment is achieved between the visual information from the images and the textual information from the questions. In the subsequent stage of \textbf{Instruction-Tuning VQA}, complex questions crafted by GPT-4V for eight specific medical scenarios are utilized to bolster the model's capacity for following instructions and understanding complex image contexts. It is worth mentioning that, unlike the methods presented in \cite{huatuogptv}, we employ the LoRA \cite{Hu2021LoRALA} technique to adjust the model parameters for better adaptation to medical scenarios.

\subsection{Uncertainty Estimation}
Uncertainty estimation, discussed as early as \cite{MacKay1992APB}, measures the predictive entropy of the output distribution $\mathbf{Y}$ given input $\mathbf{x}$ by calculating the conditional entropy $\mathbf{H}$:

\begin{equation}
\text{PE} = \mathbf{H}(\mathbf{Y} \mid \mathbf{x}) = -\sum_{\mathbf{y}} \mathbf{P}(\mathbf{y} \mid \mathbf{x}) \log \mathbf{P}(\mathbf{y} \mid \mathbf{x}).
\end{equation} High predictive entropy signals high uncertainty, with many equally likely outcomes, while low entropy indicates a concentrated output distribution. In this paper, we utilize predictive entropy to evaluate whether the answers generated by MedVLM are certain. Since generating natural language texts with LLMs or VLMs is a sequence prediction task, given a pre-trained generative MedVLM $\mathbf{P}_{\Theta}$, the log-probability of predicting the entire sequence $\mathbf{x} = \{x_1, \ldots, x_N\}$ is defined as:
\vspace{-0.2cm}

\begin{equation}
\log \mathbf{P}_{\Theta}(\mathbf{x}) = \sum_{i=1}^{N} \log \mathbf{P}_{\Theta}\left(x_i \mid \mathbf{x}_{<i}\right).
\end{equation} where \(N\) is the token lengths and the negative log-probability increases linearly with sequence length, causing longer responses to contribute more to the entropy \cite{Malinin2021UncertaintyEI}. To manage the increasing uncertainty associated with longer sequences, we normalize by the length $N$ rather than summing directly:
\vspace{-0.2cm}
\begin{equation} 
\log \mathbf{P}_{\Theta}(\mathbf{x}) = \frac{1}{N} \sum_{i=1}^{N} \log \mathbf{P}_{\Theta}\left(x_i \mid \mathbf{x}_{<i}\right).
\end{equation} The predictive entropy of the MedVLM output sequence distribution \(\mathbf{x}\) is computed as follows:

\vspace{-0.2cm}

\begin{equation}
\text{PE} = -\frac{1}{N} \sum_{i=1}^{N} \sum_{x_i} \mathbf{P}_{\Theta}\left(x_i \mid \mathbf{x}_{<i}\right) \log \mathbf{P}_{\Theta}\left(x_i \mid \mathbf{x}_{<i}\right).
\end{equation} 

For output sequences \(\mathbf{x}\) with low predictive entropy, expert intervention is bypassed, and the response is delivered directly to the user. When entropy is high, however, expert review is engaged to refine the responses. In practice, greedy search is employed to decode answers, ensuring consistent output across all iterations, with its reliability discussed in Section \ref{sec:exp}.

\subsection{Expert-Controlled Text Generation}
\textbf{Knowledge Reference Retrieval}. Although retrieval-augmented generation has shown remarkable results in natural language generation and knowledge-based question answering, its development for medical visual question answering tasks is still in its nascent stages, primarily due to two reasons: 1) the scarcity of medical multimodal knowledge bases; 2) the limited performance of medical multimodal retrievers, which struggle with the semantic gap between medical images and text. Thus, in this paper, we do not anticipate external knowledge retrieval to fully address all hallucination challenges in MedVLMs; rather, we employ it as a strategy to alleviate the burden on expert annotations. Specifically, we use BioMedCLIP \cite{BioCLIPZhang2023LargeScaleDP_49} to encode both the knowledge base and the query, employing FAISS \cite{faissJohnson2017BillionScaleSS} for k-nearest neighbors retrieval to obtain knowledge references relevant to the query. 
Given that the retrieved content may include unrelated information, we utilize the CLIP-Score for preliminary relevance assessment and automate key information extraction using the GPT-4 API \cite{GPT4TROpenAI2023}. Experts then review or edit the reference, resulting in highlighted words $\mathbf{k}$ (details in the Appendix). This context $\mathbf{k}$ is used as conditional information for classifier-free guidance \cite{Ho2022ClassifierFreeDG} to direct MedVLM’s text generation, ensuring alignment with expert-annotated key knowledge.

\textbf{Classifier-Free Guidance for MedVLM}. Classifier-Free Guidance \cite{Ho2022ClassifierFreeDG} was first introduced in conditioned Diffusion Models to generate samples aligned with the conditioning signal without relying on an explicit classifier. LLM-CFG \cite{Sanchez2023StayOT} and Prompt-Highlight \cite{Zhang2023PromptHI} later extended this concept to autoregressive language models and vision-language models, allowing for controllable text generation by adjusting the conditional prompt strength. In this paper, we apply CFG to MedVLM for generating text aligned with expert-provided knowledge. Formally, given a pre-trained MedVLM $\mathbf{P}_{\Theta}$, an input sequence $\mathbf{x}$, and expert annotation $\mathbf{k}$, the CFG sampling on MedVLM is defined as:

\begin{equation}
    \hat{\mathbf{P}}_{\Theta}(\mathbf{x} \mid \mathbf{k}) \propto \frac{\mathbf{P}_{\Theta}(\mathbf{x} \mid \mid \mathbf{k})^{\gamma}}{\mathbf{P}_{\Theta}(\mathbf{x})^{\gamma-1}} \propto \prod_{i=1}^{N} \frac{\mathbf{P}_{\Theta}\left(x_{i} \mid x_{<i},  \mathbf{k}\right)^{\gamma}}{\mathbf{P}_{\Theta}\left(x_{i} \mid x_{<i}\right)^{\gamma-1}}. 
\end{equation}Thus, the next token's logit prediction $\mathbb{P} x_{i}$ is given by:

\begin{equation}
\mathbb{P} x_{i}=\gamma \log \mathbf{P}_{\Theta}\left(x_{i} \mid x_{<i}, \mathbf{k}\right)-(\gamma-1) \log \mathbf{P}_{\Theta}\left(x_{i} \mid x_{<i}\right),
\end{equation} where $\gamma$ is the manually set guidance strength that controls the degree of generation focus. For the input context $\mathbf{x}$, a token-level binary mask \(\mathbf{m} = \{m_1, \ldots, m_N\}\) is created. This mask is used to highlight the key information $\mathbf{k}$ annotated by experts, where \(m_i = 1\) if the \(i\)-th token \(x_i\) is highlighted, and \(m_i = 0\) otherwise. This mask establishes a two-branch condition: the normal and the unconditional contexts. The normal context functions as it does in standard inference. Meanwhile, the unconditional context \(\bar{\mathbf{c}}\) is derived from the normal conditional context \(\mathbf{c} = \{c_1, \ldots, c_N\}\) within the textual embedding space through token-wise scaling:
\begin{equation}
\label{eq:alpha}
    \bar{c}_{i}=(\alpha-1) m_{i} \cdot f\left(x_{i}\right)+f\left(x_{i}\right)
\end{equation} where $\alpha$ is the scaling factor and $f(\cdot)$ is the token-to-embedding function, i.e., $c_i = f (x_i)$. We empirically set a small rescale $\alpha$ (e.g., 0.01) that can ensure a normal inference while ignoring the highlighted part. Then, based on the two-branch condition $(\mathbf{c}, \bar{\mathbf{c}})$, we can define the $i$-th token sampling process of the token-level highlight guidance as:

\begin{equation}
\label{eq:gamma}
    \begin{aligned}
\log \hat{\mathbf{P}}_{\Theta}\left(x_{i} \mid c_{<i}\right)= & \gamma \log \mathbf{P}_{\Theta}\left(x_{i} \mid c_{<i}\right) \\
& -(\gamma-1) \log \mathbf{P}_{\Theta}\left(x_{i} \mid \bar{c}_{<i}\right). 
\end{aligned}
\end{equation}

To further enhance the correlation between the tokens and the output results, we follow \cite{Zhang2023PromptHI} by adding weights at the activation score \( h_i \) in each self-attention layer of the normal context branch $\mathbf{c}$:
\vspace{-0.2cm}
\begin{equation}
\label{eq:beta}
    h_i=\log (\beta) \cdot m_i+e_i
\end{equation} where $\beta$ is the activation scaling factor, \( e_i \) represents the \( i \)-th element of the query-key multiplied attention score matrix in one attention head and the attention probability $p_i$ is calculated by:

\begin{equation}
    p_i=\frac{\exp \left(h_i\right)}{\sum_{j=1}^N \exp \left(h_j\right)}=\frac{\beta^{m_i} \cdot \exp \left(e_i\right)}{\sum_{j=1}^N \beta^{m_j} \cdot \exp \left(e_j\right)}.
\end{equation} 

For the unconditional branch $\bar{\mathbf{c}}$, the attention score is deactivated by using a scaled negative mask in the inference $\tilde{\mathbf{P}}_{\Theta}\left(x_i \mid \bar{c}_{<i}, -\delta \mathbf{m}\right)$. The hyperparameter $\gamma$ is set to 1.5, with $\alpha$ set to 0.01, and $\delta$ satisfying $\log(\beta) + 2$, which gives \(\beta = 3\). Detailed ablation studies are provided in the Appendix. This configuration enables MedVLM to generate in a controlled manner with the expert annotation token $\mathbf{k}$, without significantly impairing the model’s fundamental generative capabilities during inference.

\section{Experimental Settings}
\label{sec:exp}

\subsection{Dataset Details}
\begin{itemize}
    \item \textbf{Alignment and Instruction Tuning:} We utilize the \textbf{PubMedVision} \cite{huatuogptv} dataset, a large-scale medical VQA dataset containing 647k Alignment VQA instances and 647k Instruction Tuning VQA instances. PubMedVision extracted high-quality image-text pairs from PubMed and enhanced their quality by reformatting them using GPT-4V, significantly improving the multimodal capabilities of VLMs in the medical field.
    
    \item \textbf{Fine-tuning and Evaluation:} We evaluate our proposed method using three widely-used medical visual question answering datasets: \textbf{VQA-RAD} \cite{RADVQALau2018ADO_16} contains 3,515 question-answer pairs and 315 radiology images primarily from X-rays, CT scans, and MRIs; \textbf{SLAKE} \cite{SlakeAS_Liu2021_23} consists of 7,000 question-answer pairs and 642 images, including various modalities such as X-rays and MRIs; \textbf{PathVQA} \cite{PathVQA3QHe2020_13} includes 32,799 question-answer pairs and 4,998 pathology images from diverse histological sources. Each dataset features both open-ended and closed question types, allowing for comprehensive evaluation of our approach.
    
    \item \textbf{External Knowledge Base:} We merge all training QA pairs from the three MedVQA datasets for each image, excluding those with negative answers, to create textual descriptions for knowledge encoding. Additionally, we include two medical image-text pair datasets: \textbf{ROCO} \cite{Pelka2018RadiologyOIROCO} contains over 70,306 image-text pairs from various imaging modalities, such as ultrasound, X-rays, PET scans, CT scans, MRI, and angiography. \textbf{PEIR} \cite{wiki:peir} is a public image database for medical education, from which we extracted 33,572 image-text pairs, including microscopic and gross pathology images.
\end{itemize}

\subsection{Model Variants}
We continue alignment and fine-tuning of instructions in medical VQA datasets using the 4.2B lightweight VLMs, \textbf{Phi3-Vision} and \textbf{Phi3.5-Vision}. Both models are from the Microsoft Phi3 series \cite{Abdin2024Phi3TR}, incorporate a CLIP ViT-L/14 \cite{CLIPRadford2021LearningTV} image encoder, and are built upon the transformer architectures of Phi3-mini and Phi3.5-mini, respectively. These models undergo processes such as supervised fine-tuning and direct preference optimization, jointly training multimodal and text-only tasks so that they can achieve multimodal reasoning while maintaining language capabilities as much as possible.

\subsection{Experimental Setup and Evaluation}

We initialize the model with pre-trained weights from Phi3-Vision/Phi3.5-Vision. During training, only the LoRA layers (rank 64) were updated. For both alignment and instruction fine-tuning, we used the AdamW optimizer with a cosine learning rate scheduler, a learning rate of 2e-4, a batch size of 256, and 3 epochs. For downstream fine-tuning, we used a learning rate of 1e-4, a batch size of 64, and 10 epochs. All training was performed on 24GB NVIDIA 3090 GPUs using bfloat16 precision. Following the practice of LLaVA-Med \cite{LLaVAMedTALi2023}, we evaluate performance using accuracy for closed-set questions and recall (the ratio of ground-truth tokens appearing in the generated response) for open-set questions.

\begin{table*}[htbp]
  \centering
  \resizebox{1\textwidth}{!}{%
    \begin{tabular}{lccccccccc}
    \toprule
    \textbf{Method} &       & \textbf{VQA-RAD} &       &       & \textbf{SLAKE} &       &       & \textbf{PathVQA} &  \\
    \cmidrule{2-10}          & Open  & Closed & Overall & Open  & Closed & Overall & Open  & Closed & Overall \\
    \midrule
    \multicolumn{10}{l}{\textit{Classification-based MedVQA methods}} \\
    M2I2 \cite{M2I2Li2022SelfsupervisedVP_22}  & 66.50 & 83.50 & 76.80 & 74.70 & 91.10 & 81.20 & 36.30 & 88.00 & 62.20 \\
    Prefix \cite{PrefixSonsbeek2023OpenEndedMV_41} & -     & -     & -     & 84.30 & 82.01 & 83.30 & 40.00 & 87.00 & 63.60 \\
    PubMedCLIP \cite{PubMedCLIPHMEslami2023_8} & 60.10 & 80.00 & 72.10 & 78.40 & 82.50 & 80.10 & -     & -     & - \\
    M3AE \cite{Bai2024M3DA3}  & 67.23 & 83.46 & 77.01 & 80.31 & 87.82 & 83.14 & -     & -     & - \\
    BioMedCLIP \cite{BioCLIPZhang2023LargeScaleDP_49} & 67.60 & 79.80 & 75.20 & 82.50 & 89.70 & 85.40 & -     & -     & - \\
    MUMC \cite{Li2023MaskedVA}  & 71.50 & 84.20 & 79.20 & -     & -     & 84.90 & 39.00 & 90.40 & 65.10 \\
    \midrule
    \multicolumn{10}{l}{\textit{Autoregressive-based MedVQA methods}} \\
    LLaVA-Med 7B \cite{LLaVAMedTALi2023} & 61.52 & 84.19 & 75.19 & 83.08 & 85.34 & 83.97 & 37.95 & 91.21 & 64.66 \\
    LLaVA-Med 13B \cite{LLaVAMedTALi2023} & 64.58 & 77.97 & 72.66 & 84.97 & 85.58 & 85.21 & 38.82 & 92.39 & 65.69 \\
    BioMed-VITAL (13B, 60K) \cite{BioMedVITAL} & 64.88 & 84.55 & 76.74 & \underline{87.82} & 86.54 & 87.32 & 39.71 & 91.41 & 65.64 \\
    BioMed-VITAL (13B, 150K) \cite{BioMedVITAL} & 69.72 & 84.86 & 78.85 & \textbf{91.69} & \underline{90.70} & \textbf{91.30} & 39.89 & 92.42 & 66.24 \\
    \midrule
    \multicolumn{10}{l}{\textit{Our supervised fine-tuning results from Phi3V / Phi3.5V-based model}} \\
    Phi3V-Med 4.2B & 70.39 & 85.29 & 79.38 & 82.02 & 87.02 & 83.98 & 36.88 & 91.80 & 64.43 \\
    
    \quad w/ RAG & 66.48 (\textcolor{green}{-3.91}) & 90.07 (\textcolor{red}{+4.78}) & 80.71 (\textcolor{red}{+1.33}) & 83.41 (\textcolor{red}{+1.39}) & 87.74 (\textcolor{red}{+0.72}) & 85.11 (\textcolor{red}{+1.13}) & \underline{49.05} (\textcolor{red}{+12.17}) & \textbf{95.28} (\textcolor{red}{+3.48}) & \underline{72.24} (\textcolor{red}{+7.81}) \\

    \quad w/ 5\% RAG & 68.72 (\textcolor{green}{-1.67}) & 85.29 (\textcolor{red}{+0.00}) & 78.71 (\textcolor{green}{-0.67}) & 82.33 (\textcolor{red}{+0.31}) & 87.50 (\textcolor{red}{+0.48}) & 84.35 (\textcolor{red}{+0.37}) & 38.81 (\textcolor{red}{+1.93}) & 91.80 (\textcolor{red}{+0.00}) & 65.39 (\textcolor{red}{+0.96}) \\

    \quad w/ 5\% Expert-RAG & 71.50 (\textcolor{red}{+1.11}) & 86.03 (\textcolor{red}{+0.74}) & 80.27 (\textcolor{red}{+0.89}) & 83.25 (\textcolor{red}{+1.23}) & 88.46 (\textcolor{red}{+1.44}) & 85.29 (\textcolor{red}{+1.31}) & 42.37 (\textcolor{red}{+5.49}) & 91.80 (\textcolor{green}{0.00}) & 67.16 (\textcolor{red}{+2.73}) \\

    \rowcolor[rgb]{ .749,  .749,  .749} \quad w/ 5\% Expert-CFG & 75.42 (\textcolor{red}{+5.03}) & 86.03 (\textcolor{red}{+0.74}) & 81.82 (\textcolor{red}{+2.44}) & 84.96 (\textcolor{red}{+2.94}) & 88.46 (\textcolor{red}{+1.44}) & 86.33 (\textcolor{red}{+2.35}) & 45.88 (\textcolor{red}{+9.00}) & 91.80 (\textcolor{red}{+0.00}) & 68.91 (\textcolor{red}{+4.48}) \\

    Phi3.5V-Med 4.2B & 69.27 & \underline{90.44} & 82.04 & 82.95 & 88.70 & 85.20 & 37.42 & 92.89 & 65.24 \\

    \quad w/ RAG & 70.39 (\textcolor{red}{+1.12}) & 88.24 (\textcolor{green}{-2.20}) & 81.15 (\textcolor{green}{-0.89}) & 82.79 (\textcolor{green}{-0.16}) & 89.90 (\textcolor{red}{+1.20}) & 85.58 (\textcolor{red}{+0.38}) & \textbf{51.34} (\textcolor{red}{+13.92}) & \underline{94.34} (\textcolor{red}{+1.45}) & \textbf{72.90} (\textcolor{red}{+7.66}) \\

    \quad w/ 5\% RAG & 70.95 (\textcolor{red}{+1.68}) & 89.34 (\textcolor{green}{-1.10}) & 82.04 (\textcolor{red}{+0.00}) & 82.64 (\textcolor{green}{-0.31}) & 89.42 (\textcolor{red}{+0.72}) & 85.30 (\textcolor{red}{+0.10}) & 39.02 (\textcolor{red}{+1.60}) & 92.89 (\textcolor{red}{+0.00}) & 66.04 (\textcolor{red}{+0.80}) \\

    \quad w/ 5\% Expert-RAG & 72.06 (\textcolor{red}{+2.79}) & \underline{90.44} (\textcolor{red}{+0.00}) & \underline{83.37} (\textcolor{red}{+1.33}) & 83.56 (\textcolor{red}{+0.61}) & 89.90 (\textcolor{red}{+1.20}) & 86.05 (\textcolor{red}{+0.85}) & 41.72 (\textcolor{red}{+4.30}) & 92.74 (\textcolor{green}{-0.15}) & 67.31 (\textcolor{red}{+2.07}) \\

    \rowcolor[rgb]{ .749,  .749,  .749} \quad w/ 5\% Expert-CFG & \underline{74.86} (\textcolor{red}{+5.59}) & \textbf{91.18} (\textcolor{red}{+0.74}) & \textbf{84.70} (\textcolor{red}{+2.66}) & 85.27 (\textcolor{red}{+2.32}) & \textbf{91.59} (\textcolor{red}{+2.89}) & \underline{87.75} (\textcolor{red}{+2.55}) & 46.08 (\textcolor{red}{+8.66}) & 92.89 (\textcolor{red}{+0.00}) & 69.56 (\textcolor{red}{+4.32}) \\
    
    \bottomrule
    \end{tabular}%
  } 
  \caption{\textbf{Performance comparison of MedVQA methods across three datasets.} The evaluated methods include representative state-of-the-art classification-based and autoregressive-based methods, as well as our proposed \textbf{Expert-CFG} based on Phi3V and Phi3.5V. \textit{w/ RAG} represents retrieval for the top-1 match on all queries, while \textit{w/ 5\% RAG} applies retrieval to the top 5\% high-entropy predictions. \textit{Expert-RAG} and \textit{Expert-CFG} add expert annotations, with \textit{Expert-CFG} using classifier-free guidance for expert-aligned output.}
  \label{tab:sota}
\end{table*}

\section{Experimental Results}
\subsection{Compared with SoTAs}
Table \ref{tab:sota} summarizes the performance metrics of \textbf{Expert-CFG} compared with models based on Phi3V and Phi3.5V, as well as the reported results of representative state-of-the-art (SoTA) methods. We divide existing methods into two categories: 1) those that treat open-ended question-answering tasks as classification among answer candidates in the training set; and 2) those based on autoregressive open-ended questions. Although the classification-based MedVQA achieves quite competitive accuracy, it does not meet the practical application needed for open-ended question answering. Table \ref{tab:sota} shows that our baseline model surpasses some 7B and 13B models on the VQA benchmark thanks to high-quality medical visual instruction data \cite{huatuogptv} and the powerful Phi3V model \cite{Abdin2024Phi3TR}. To quantitatively evaluate the effectiveness of the proposed \textbf{Expert-CFG}, we present the results of \textit{RAG}, \textit{RAG on the top 5\% high-entropy samples}, and the expert-enhanced methods \textit{Expert-RAG} and \textit{Expert-CFG} (both applied on the top 5\% high-entropy samples). As shown in the lower part of Table \ref{tab:sota}, \textit{RAG} demonstrates limited performance improvements due to inaccuracies in retrieved knowledge \cite{Xia2024RULERM, Xia2024MMedRAGVM}. Although expert annotations were introduced as a gold standard, the effectiveness of \textit{Expert-RAG} is significantly lower than that of \textit{Expert-CFG} due to potential conflicts between retrieved knowledge and the model’s internal knowledge. We discuss these limitations in the ablation study section \ref{sec:abl}.
\subsection{Ablation Study}
\label{sec:abl}
To explore the rationality and effectiveness of each component, we primarily investigate the following three research questions in this chapter: \textbf{1) Can entropy accurately assess model reliability? 2) What are the limitations of retrieval augmentation in medical field? 3) How to balance expert workload and accuracy?} To ensure a fair comparison, all experiments are conducted based on a baseline constructed with Phi3.5V and PubMedVision.

\begin{figure}[htbp]
    \centering
    \begin{minipage}[t]{0.49\columnwidth}
        \centering
        \includegraphics[width=\linewidth]{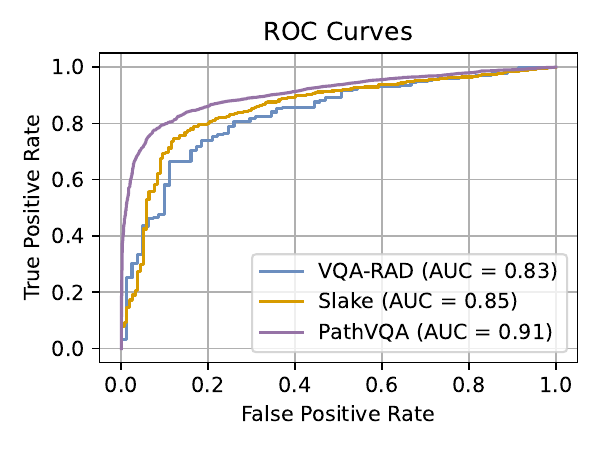}
        \subcaption{} 
        \label{fig:sub_roc_auc}
    \end{minipage}
    \hfill 
    \begin{minipage}[t]{0.49\columnwidth}
        \centering
        \includegraphics[width=\linewidth]{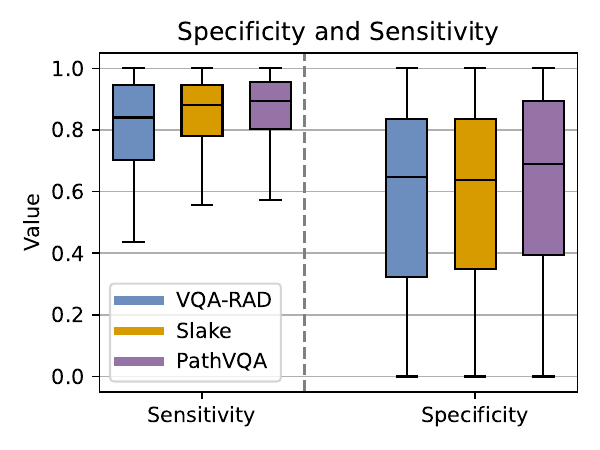}
        \subcaption{}
        \label{fig:sub_box_plot} 
    \end{minipage}
    \caption{\textbf{Uncertainty Estimation Performance of Three MedVQA Benchmarks.} (a) Receiver Operating Characteristic (ROC) curves and corresponding Area Under the Curve (AUC). (b) Box plots of sensitivity and specificity.}
    \label{fig:uncertainty}
\end{figure}

\textbf{Uncertainty Estimation Performance}. We assess uncertainty estimation efficacy on three MedVQA benchmarks featuring short-answer tasks. Given the prohibitive computational costs of multiple sampling and the constrained combinatorial space inherent to concise answers, our Expert-CFG pipeline employs greedy decoding-based prediction entropy for uncertainty quantification. After normalizing entropy values and applying threshold-based classification, we find that high entropy strongly correlates with incorrect predictions. The performance metric, AUC (Area Under the Curve, used to distinguish between correct and incorrect answers), exceeds 0.8 across all datasets, as shown in Figure \ref{fig:uncertainty}. This demonstrates the ability of our method to distinguish between positive and negative samples. Notably, elevated entropy often results from sparse training samples or ambiguous annotations. For a more detailed analysis of the uncertainty estimation methods and the entropy across different answer error types, please refer to the Appendix.

\begin{figure*}[htbp]
    \centering
    \captionsetup{skip=0pt}
    \begin{minipage}[c]{\textwidth}
        \centering
        \resizebox{\textwidth}{!}{%
            \includegraphics{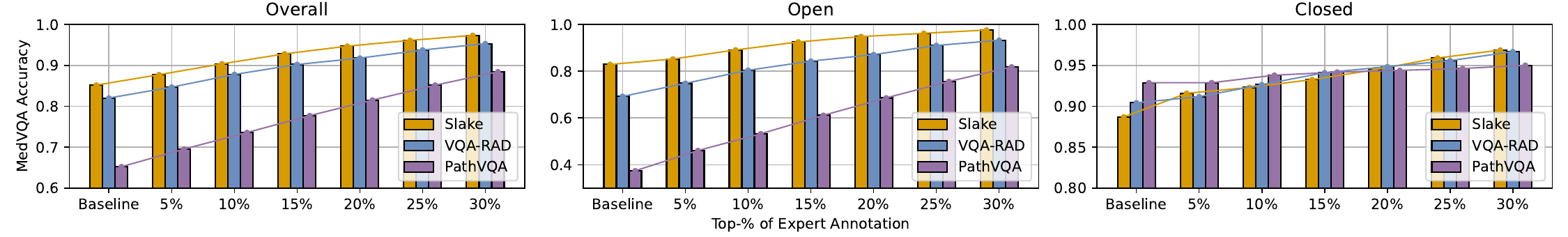}
        }
        \subcaption{Varying percentages of high-entropy samples from the test set.}
        \label{fig:topk_expanno}
    \end{minipage}
    \vspace{1em}
    \begin{minipage}[c]{\textwidth}
        \centering
        \resizebox{\textwidth}{!}{%
            \includegraphics{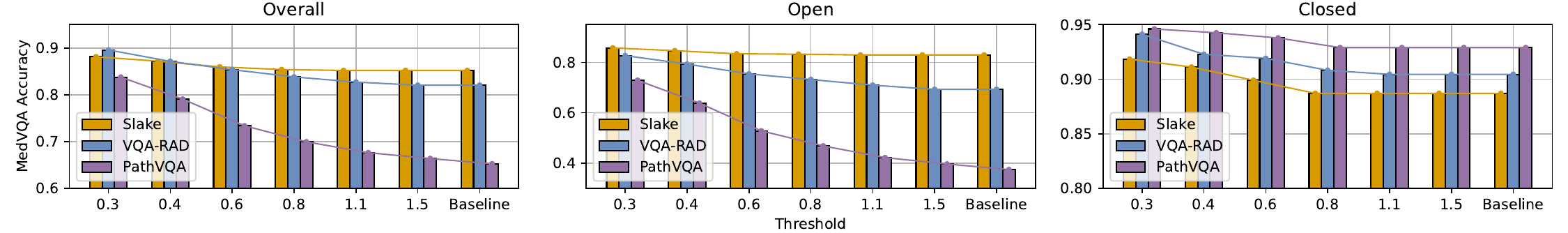}
        }
        \subcaption{Fixed entropy threshold.}
        \label{fig:thresh_expanno}
    \end{minipage}
    \vspace{-0.2cm}
    \caption{\textbf{MedVQA accuracy across three datasets with additional expert annotations.} Note that expert annotations (highlights) are generated automatically through entity extraction and string matching.}
\end{figure*}

\begin{figure}[H]
    \centering
    \resizebox{1\columnwidth}{!}{%
        \includegraphics{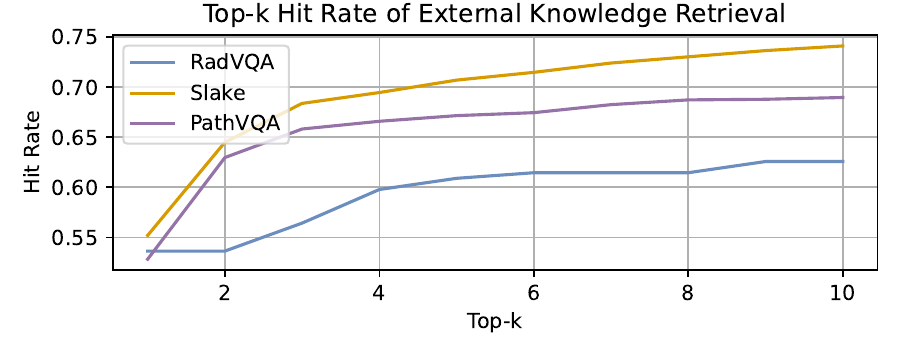}
    }
    \vspace{-0.7cm}
    \caption{\textbf{Hit Rate of answers directly appearing in retrieved captions for Top-K retrieval.} Considering that the answers to closed-ended questions, such as ``yes" and ``no," do not directly appear in captions, we evaluate only open-ended questions.}
    \label{fig:rag_hitrate}
\end{figure}

\begin{table}[H]
  \centering
  \resizebox{0.8\columnwidth}{!}{%
    \begin{tabular}{lccc}
    \toprule
    \textbf{Feature Type} & \textbf{VQA-RAD} & \textbf{SLAKE} & \textbf{PathVQA} \\
    \midrule
    Image & 53.07  & 47.75  & 51.69  \\
    Text  & 53.07  & 54.57  & 46.14  \\
    Sum   & 53.07  & 54.26  & 47.83  \\
    \rowcolor[rgb]{ .749,  .749,  .749} \textbf{Union} & 53.07  & \textbf{55.97}  & \textbf{52.70}  \\
    \bottomrule
    \end{tabular}%
  } 
  \caption{\textbf{Top-1 Hit Rate across different feature types.} ``Sum" uses the combined image and text features, while ``Union" retrieves based on each feature separately, selecting the result with the highest similarity.}
  \label{tab:rag_method}
\end{table}

\textbf{Retrieval Performance}. Although we observed that retrieval-augmented generation (RAG) alone provides limited improvement (see Table \ref{tab:sota}), the complexity of medical data makes it difficult to determine whether the root cause lies in the knowledge base or the retriever. To investigate this, we conducted ablation experiments on the combined external medical knowledge base, consisting of over 100,000 image-text pairs. We directly assess whether the retrieved information aids answer generation by using the answer hit rate, as shown in Figure \ref{fig:rag_hitrate} and Table \ref{tab:rag_method}. Despite varying retrieval approaches, the Top-1 retrieval hit rate for all three datasets does not exceed 56\%, whether using single-modality retrieval or combined image-text feature retrieval. The hit rates for VQA-RAD are identical across different feature types. Furthermore, we observe that when the number of retrieved entries \( K \) increases to 4, the hit rate reaches a bottleneck, echoing similar findings reported in RAMM \cite{Yuan2023RAMMRB}.

\begin{figure*}[htbp]
\centering
\includegraphics[width=0.98\textwidth]{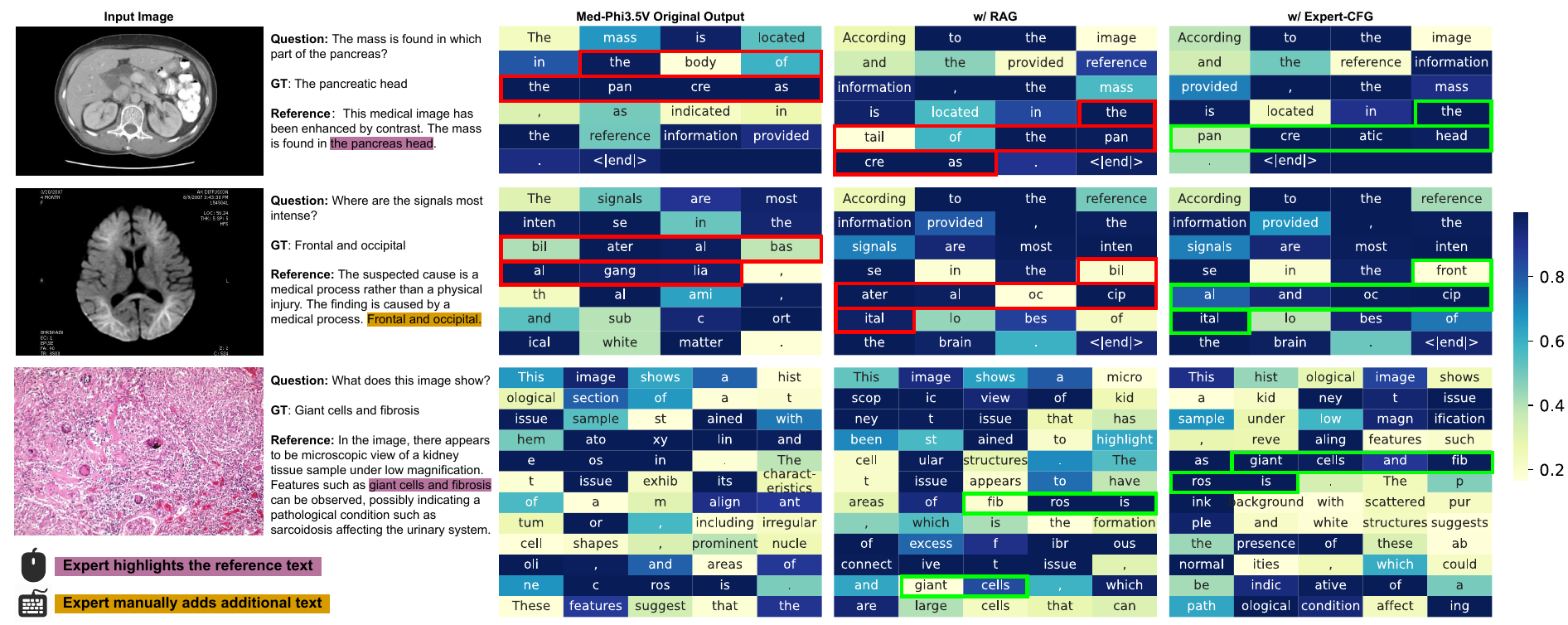}
\caption{\textbf{Examples from VQA-RAD, SLAKE, and PathVQA datasets for zero-shot predictions by Phi3.5V-Med}. The probability of each token is visualized using a colormap, comparing the original output, retrieval-augmented generation (RAG) with expert annotations, and expert-controlled classifier-free guidance (Expert-CFG).  }
\label{fig:case}
\end{figure*}

\textbf{Balancing Expert Workload and Accuracy}. The introduction of expert annotations equivalent to gold standards undoubtedly enhances the performance of MedVLM. However, it is impractical to require experts to verify each output of MedVLM individually, as this would result in an enormous workload. Therefore, we employ two discrimination methods: one based on data percentage (selecting the top percentage of high-entropy samples) and another based on a fixed entropy threshold. Based on the uncertainty level of the answers, we require experts to annotate a certain number of samples and conduct quantitative evaluations on three MedVQA benchmarks. To reduce the workload, for open-ended questions, we directly use the ground truth as expert annotations. For closed-ended questions where the answers are only ``yes" or ``no" and provide no additional information, we introduce retrieved texts and use GPT-4 API \cite{GPT4TROpenAI2023} for entity extraction, matching them with the keywords in the questions to highlight relevant information, serving as expert annotations. For the few datasets that lack such annotations, manual annotation is performed. The relationship between the answer accuracy percentage and entropy threshold for the three datasets is shown in Figure \ref{fig:topk_expanno} and \ref{fig:thresh_expanno}. It can be seen that although accuracy is proportional to the number of expert annotations, the improvement eventually reaches a bottleneck. Considering the diminishing marginal returns, we suggest focusing expert annotations on the top 5\% of high-entropy samples to achieve a balance between workload and accuracy.

\begin{table}[H]
  \centering
  \resizebox{0.85\columnwidth}{!}{%
    \begin{tabular}{lccc}
    \toprule
    Model Variants \& Data & \multicolumn{1}{c}{\textbf{VQA-RAD}} & \multicolumn{1}{c}{\textbf{SLAKE}} & \multicolumn{1}{c}{\textbf{PathVQA}} \\
    \midrule
    LLaVA 7B  & 59.19 & 50.24 & 45.65 \\
    \quad +LLaVA-Med  \cite{LLaVAMedTALi2023} & 60.66 & 54.33 & 49.07 \\ \midrule
    Phi3V 4.2B & 54.98 & 55.77 & 52.28 \\
     \quad +LLaVA-Med  \cite{LLaVAMedTALi2023} & 70.96 & 60.58 & 53.94 \\
     \quad +PubMedVision \cite{huatuogptv} & 73.90 & 67.31 & 54.38 \\
    \midrule
    Phi3.5V 4.2B & 57.72 & 57.93 & 53.52 \\
     \quad +PubMedVision \cite{huatuogptv} & \textbf{74.63} & \textbf{69.08} & \textbf{54.79} \\
    \bottomrule
    \end{tabular}%
  }
  \vspace{-0.2cm}
  \caption{\textbf{Zero-Shot Performance comparison of model variants and instruction dataset.} Evaluated only on closed QA.}
  \label{tab:model}%
\end{table}%

\textbf{Model Variants and Instruction Data}. Although this is not the main focus of this paper, a powerful MedVLM is crucial for building a reliable medical visual question-answering process. Due to GPU memory limitations, we abandoned LLaVA \cite{llavaLiu2023VisualIT} and instead focused on the smaller 4.2B Phi3 series models, which have competitive performance on public evaluation benchmarks. Specifically, we experimented with different variants of the Phi models, such as Phi3V and Phi3.5V, to assess their performance in medical VQA tasks. Considering that the baseline model has limited parameters, we can only rely on high-quality medical instruction alignment. To this end, we experimented with all existing open-source medical instruction fine-tuning datasets, including LLaVA-Med \cite{LLaVAMedTALi2023} and PubMedVision \cite{huatuogptv}. Table \ref{tab:model} shows experimental results, highlighting performance gains from high-quality data and a strong baseline model.

\subsection{Case Study}

Figure \ref{fig:case} illustrates the performance of Phi3.5V-Med in answering questions related to untrained medical images. For the abdominal CT and brain MRI, the Phi3.5V-Med model fine-tuned with PubMedVision produced incorrect answers regarding specific details. For the microscopic image, although the generated answer contained no obvious errors, it failed to capture key information. Next, retrieval-augmented generation (w/ RAG) was used to provide additional knowledge references. Specifically, for the brain MRI, as the correct answer was missing in the reference, an expert manually added the key information. However, Phi3.5V-Med did not fully follow the provided reference text until the Expert-CFG was applied, resulting in a complete and correct answer. Notably, compared to Expert-CFG, the original model and the model with RAG both showed lower confidence in tokens related to key information, as represented by lighter colors. In contrast, Expert-CFG exhibited relatively higher confidence, represented by darker colors.

\section{Conclusion}

In this paper, we propose Expert-CFG, a novel framework that directly aligns MedVLMs with clinical expertise through low-cost annotations, enhancing their medical vision-language performance. Experimental results demonstrate that our framework outperforms existing approaches, achieving state-of-the-art performance on three medical visual question-answering benchmarks. Future work will focus on refining the integration of experts or expert models and exploring scalability to other medical applications, particularly in addressing knowledge gaps and long-tail challenges in MedVLM datasets.

\section*{Acknowledge}

This work was supported in part by the National Science and Technology Major Project under Grant 2022ZD0117103, in part by the Outstanding Youth Science Foundation of Shaanxi Province under Grant 2025JC-JCQN-083, in part by the National Natural Science Foundation of China under Grant 62302354, in part by the Key Research and Development Program of Shaanxi Province under Grant 2025CY-YBXM-047, and in part by the Fundamental Research Funds for the Central Universities under Grant QTZX23084.
{
    \small
    \bibliographystyle{ieeenat_fullname}
    \bibliography{main}
}
\appendix
\onecolumn

\section{Dataset Details}

\begin{table*}[h]
  \centering
  \resizebox{1\textwidth}{!}{
    \begin{tabular}{l|ccccc}
    \toprule
    \textbf{Dataset} & \multicolumn{1}{c}{\textbf{Image}} & \multicolumn{1}{c}{\textbf{Description}} & \textbf{QA-Pairs} & \textbf{Type} & \textbf{Usage} \\
    \midrule
    VQA-RAD \cite{RADVQALau2018ADO_16} & 315    & 315    & \multicolumn{1}{c}{3,515}  & Radiology & Downstream Fine-tuning \& Evaluation \& RAG \\
    SLAKE \cite{SlakeAS_Liu2021_23}   & 642    & 642    & \multicolumn{1}{c}{7,000}  & Radiology & Downstream Fine-tuning \& Evaluation \& RAG \\
    PathVQA \cite{PathVQA3QHe2020_13} & 4,998  & 4,998  & \multicolumn{1}{c}{32,799} & Pathology & Downstream Fine-tuning \& Evaluation \& RAG \\
    ROCO \cite{Pelka2018RadiologyOIROCO}    & 70,306 & 70,306 & \multicolumn{1}{c}{—}      & Radiology & Retrieval Augmentation \\
    PEIR \cite{wiki:peir}    & 33,572 & 33,572 & \multicolumn{1}{c}{—}      & Pathology & Retrieval Augmentation \\
    PubMedVision \cite{huatuogptv} & 1,009,700 & 647,031 & \multicolumn{1}{c}{647,031} & Comprehensive & Alignment \& Instruct Tuning \\
    \bottomrule
    \end{tabular}%
  }
  \caption{\textbf{Dataset statistics}, including the number of images, image descriptions, QA-pairs, and data types are detailed. For MedVQA data, only the \textbf{training set} is used for knowledge retrieval.}
  \label{tab:dataset}
\end{table*}

\section{Prompting GPT-4 to Convert Question and Answer into Caption}
\begin{figure*}[h]
    \centering
    \includegraphics[width=0.7\linewidth]{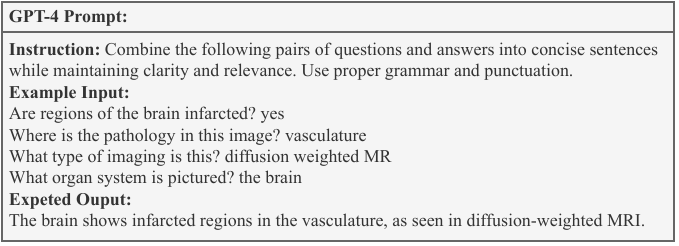}
    \caption{\textbf{GPT-4 prompt for converting question-answer pairs into concise medical captions.}}
    \label{fig:prompt_caption}
\end{figure*}

Considering that BioMedCLIP \cite{BioCLIPZhang2023LargeScaleDP_49} is trained on paired medical images and textual captions, we reformulated the question-answer pairs in the MedVQA dataset into concise captions to ensure semantic consistency. Specifically, questions with answers like ``no'' were excluded, as such negative responses provide no meaningful information for generating captions. Figure \ref{fig:prompt_caption} illustrates the GPT-4 prompt used for converting question-answer pairs into coherent and clinically relevant captions.

\section{Prompting GPT-4 to Generate Expert Annotations}
\begin{figure*}[htbp]
    \centering
    \includegraphics[width=0.7\linewidth]{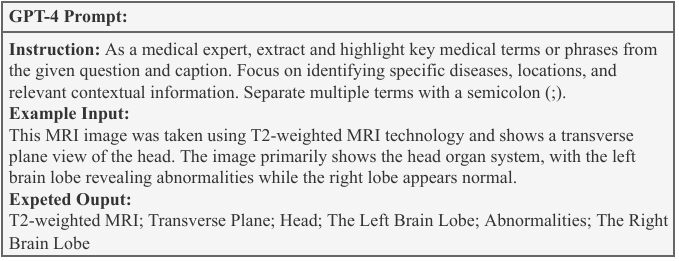}
    \caption{\textbf{GPT-4 prompt for extracting medical terms from captions.}}
    \label{fig:prompt_keyword}
\end{figure*}

\begin{figure*}[h]
\centering
\includegraphics[width=0.7\textwidth]{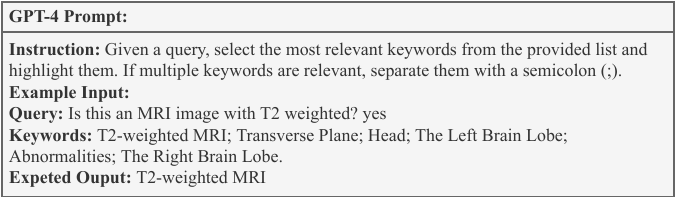}
\caption{\textbf{GPT-4 prompt for highlighting relevant key medical terms based on a query.}}
\label{fig:prompt_highlight}
\end{figure*}

Expert-CFG requires experts to predefine semantically coherent and query-relevant keywords or keyword groups as highlights. However, relying entirely on experts for this task is clearly impractical. To address this, we propose a semi-automated pipeline that combines expert guidance with automated keyword extraction techniques to streamline the process. To further optimize token usage, we divide the process into two steps: 1) extracting keywords from each caption in the knowledge base (see Figure \ref{fig:prompt_keyword}); 2) matching these extracted keywords with the query to identify the most relevant highlights (see Figure \ref{fig:prompt_highlight}). Since there is no keyword annotation in the existing MedVQA, we only preliminarily evaluated the performance of this method on the VQA-RAD dataset with the smallest amount of data, as shown in Table \label{tab:keyword_extraction}. A recall rate of over 98\% indicates that GPT-4o can extract keywords from captions well. 

\begin{table}[htbp]
    \centering
        \begin{tabular}{lccc}
            \toprule
            Method & \multicolumn{3}{c}{Keyword Extraction on VQA-RAD} \\
            \cmidrule{2 - 4}
            & Recall & Precision & F1 \\
            \midrule
            GPT4o & 89.57 & 98.36 & 93.76 \\
            \bottomrule
        \end{tabular}
    \caption{Performance of GPT-4o in extracting keyword from captions.}
    \label{tab:keyword_extraction}
\end{table}

\section{Discussion on Uncertainty Estimation Metric}

For uncertainty estimation, we selected the entropy of the generated answer. This choice was motivated by a combination of effectiveness, efficiency, and system compatibility. Specifically, entropy proved to be a reliable uncertainty indicator for our model and integrates seamlessly with our Expert-CFG framework. A key advantage is its computational efficiency: the entropy can be calculated from a single forward pass of our VLM without requiring any architectural modifications or additional inference costs. We considered alternative methods, such as those based on multiple forward passes (e.g., Monte Carlo dropout). However, these were deemed unsuitable as they introduce significant computational overhead, conflicting with our design goals. Our approach aligns with recent findings that favor direct, single-pass uncertainty measures for large language models \cite{Tian2023JustAF}.

\begin{table*}[htbp]
  \centering
    \resizebox{\textwidth}{!}{
    \begin{tabular}{cl|rrrr|rrrr|r}
    \toprule
          &       & \multicolumn{4}{c|}{\textbf{VQA-RAD(Closed)}} & \multicolumn{4}{c|}{\textbf{SLAKE(Closed)}} & \multicolumn{1}{p{11.82em}}{\textbf{Implementation }} \\
\cmidrule{3-10}          & \textbf{Method} & \multicolumn{1}{c}{\textbf{ECE$\downarrow$}} & \multicolumn{1}{c}{\textbf{ECE-t$\downarrow$}} & \multicolumn{1}{c}{\textbf{BS-t$\downarrow$}} & \multicolumn{1}{c|}{\textbf{AUC$\uparrow$}} & \multicolumn{1}{c}{\textbf{ECE$\downarrow$}} & \multicolumn{1}{c}{\textbf{ECE-t$\downarrow$}} & \multicolumn{1}{c}{\textbf{BS-t$\downarrow$}} & \multicolumn{1}{c|}{\textbf{AUC$\uparrow$}} & \multicolumn{1}{p{11.82em}}{\textbf{Details \& Explanation}} \\
    \midrule
    \multicolumn{1}{c}{\multirow{3}[2]{*}{\makecell{Phi3.5V-Med\\Zero-Shot}}} & Label prob. & 0.333 & 0.042 & 0.230 & 0.636 & 0.117 & 0.039 & 0.196 & 0.691 & \multicolumn{1}{p{11.82em}}{do sample=10, temp=1} \\
          & Is True prob. & 0.174 & 0.095 & 0.201 & 0.727 & 0.294 & 0.094 & 0.212 & 0.617 & \multicolumn{1}{p{11.82em}}{$\mathcal{Q}+\mathcal{A}$ ``\textit{is True?}''} \\
          & Entropy & 0.108 & 0.037 & 0.145 & 0.879 & 0.119 & 0.066 & 0.148 & 0.827 & \multicolumn{1}{p{11.82em}}{$\mathcal{A}$ token prob (greedy)} \\
    \midrule
    \multicolumn{1}{c}{\multirow{2}[2]{*}{\makecell{Phi3.5V-Med\\Fine-Tune}}} & Label prob. & 0.089 & 0.019 & 0.077 & 0.889 & 0.082 & 0.026 & 0.071 & 0.895 & \multicolumn{1}{p{11.82em}}{This model can not follow} \\
          & Entropy & 0.068 & 0.024 & 0.066 & 0.958 & 0.073 & 0.024 & 0.063 & 0.949 & \multicolumn{1}{p{11.82em}}{ ``\textit{is True}'' prompt} \\
    \midrule
    \multicolumn{1}{c}{\multirow{4}[2]{*}{\makecell{HuatuoV-32B\\Zero-Shot}}} & Label prob. & 0.266 & 0.200 & 0.231 & 0.811 & 0.190 & 0.056 & 0.154 & 0.768 &  \\
          & Is True prob. & 0.114 & 0.116 & 0.172 & 0.828 & 0.082 & 0.074 & 0.151 & 0.851 & \multicolumn{1}{p{11.82em}}{Stronger instruct following.} \\
          & Verb1s top-1 & 0.166 & 0.001 & 0.166 & 0.754 & 0.141 & 0.053 & 0.145 & 0.808 &  \multicolumn{1}{p{11.82em}}{Fixed outputs(e.g., 0.95).} \\
          & Entropy & 0.041 & 0.044 & 0.141 & 0.886 & 0.055 & 0.048 & 0.126 & 0.885 &  \\
    \bottomrule
    \end{tabular}
    }
    \caption{Uncertainty metrics on VQA-RAD / SLAKE (``\textit{yes/no}'' QA) under zero-shot / fine-tuned settings. }
  \label{tab:uncertainty_metrics_rad_slake}
\end{table*}

\section{Entropy and Retrieval Hit Rate of Answers}

\begin{figure}[h]
    \centering
    \begin{minipage}{1\textwidth}
        \centering
        \includegraphics[width=\textwidth]{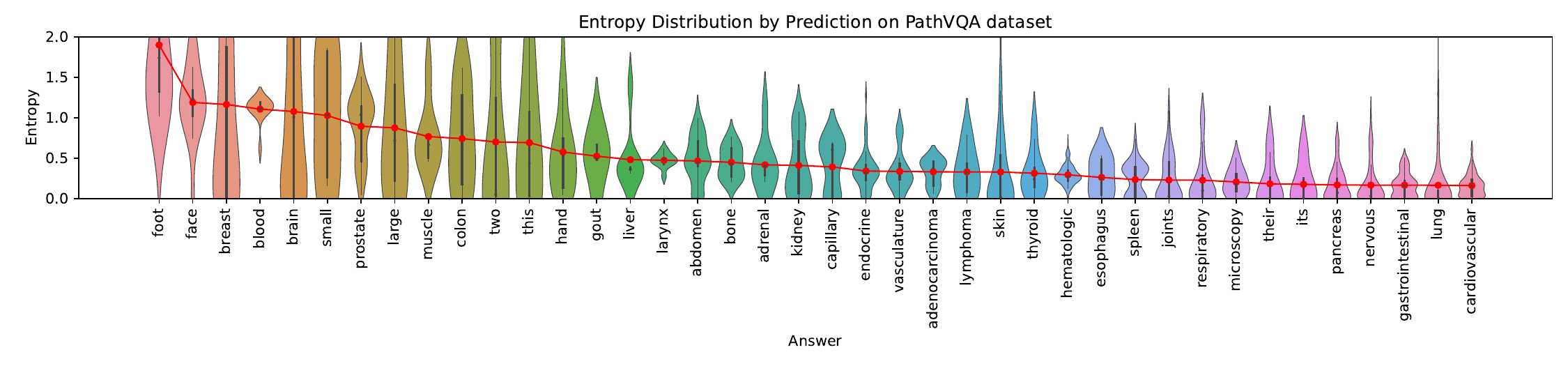}
        \subcaption{}
    \end{minipage}%
    \hfill
    \begin{minipage}{1\textwidth}
        \centering
        \includegraphics[width=\textwidth]{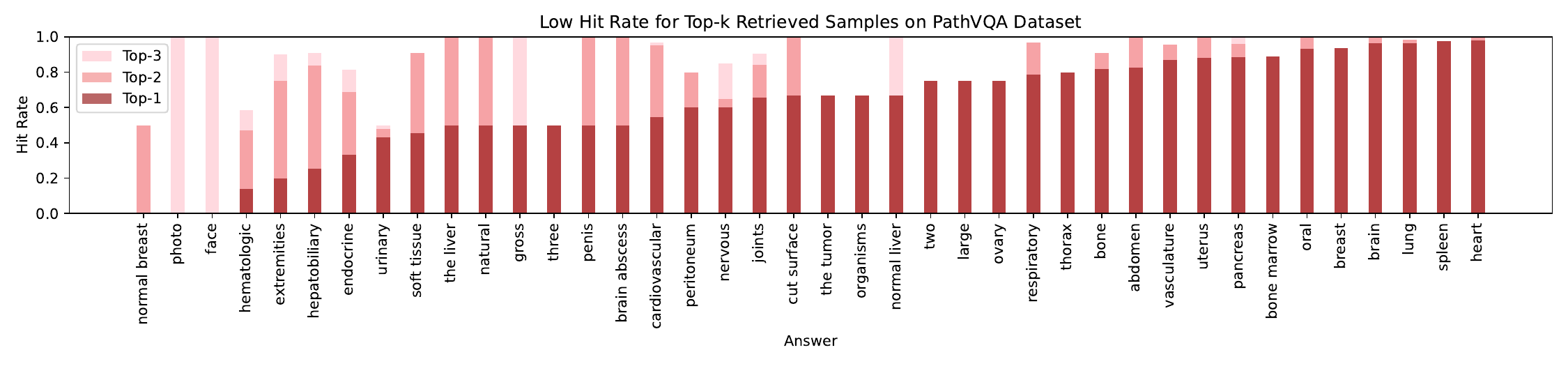}
        \subcaption{}
    \end{minipage}
    \caption{\textbf{PathVQA Dataset}: (a) Entropy Distribution and (b) Overlapping Answer Hit Rate for Top-k.}
    \label{fig:path_distribution_hit_rate}
\end{figure}

\begin{figure}[H]
    \centering
    \begin{minipage}{0.5\textwidth}
        \centering
        \includegraphics[width=\textwidth]{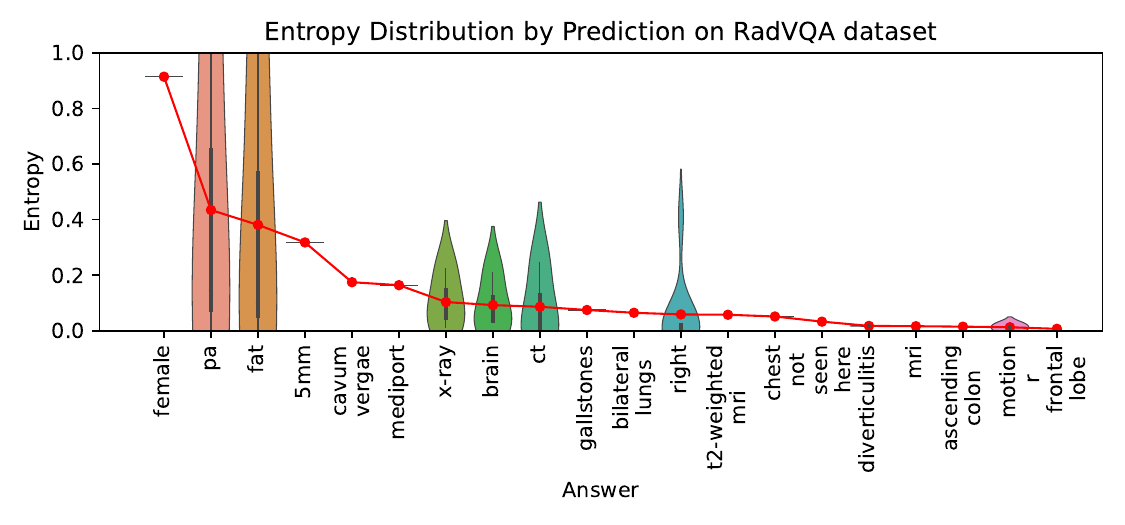}
        \subcaption{}
    \end{minipage}%
    \hfill
    \begin{minipage}{0.5\textwidth}
        \centering
        \includegraphics[width=\textwidth]{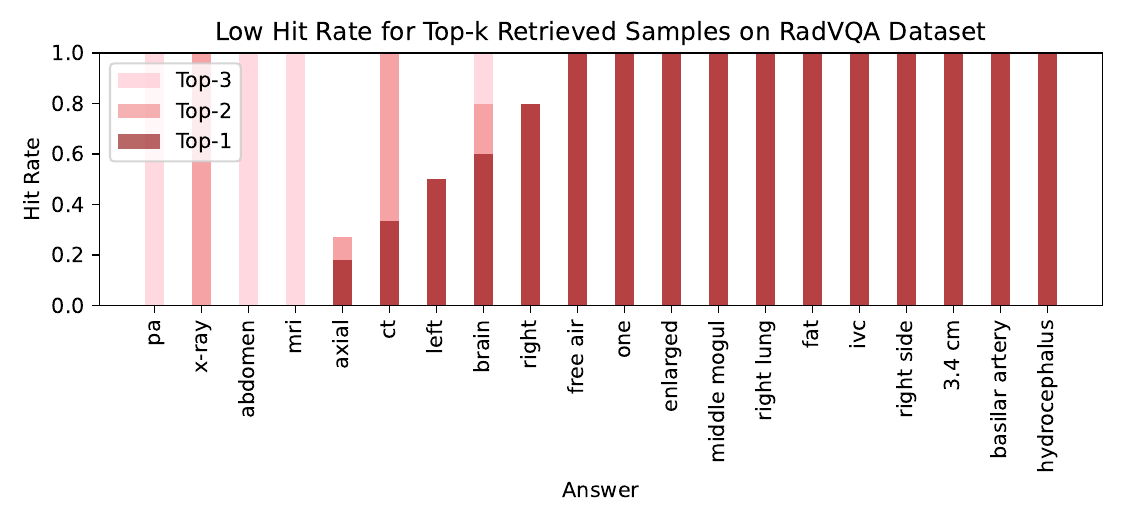}
        \subcaption{}
    \end{minipage}
    \caption{\textbf{VQA-RAD Dataset}: (a) Entropy Distribution and (b) Overlapping Answer Hit Rate for Top-k.}
    \label{fig:rad_distribution_hit_rate}
\end{figure}

\begin{figure}[H]
    \centering
    \begin{minipage}{0.8\textwidth}
        \centering
        \includegraphics[width=\textwidth]{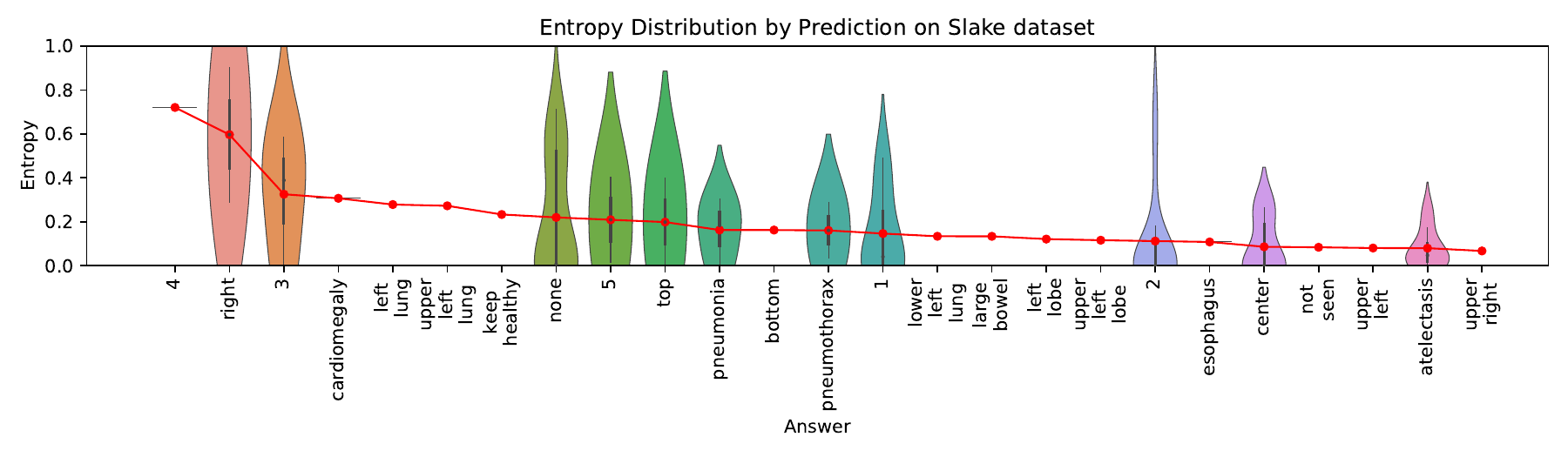}
        \subcaption{}
    \end{minipage}%
    \hfill
    \begin{minipage}{0.8\textwidth}
        \centering
        \includegraphics[width=\textwidth]{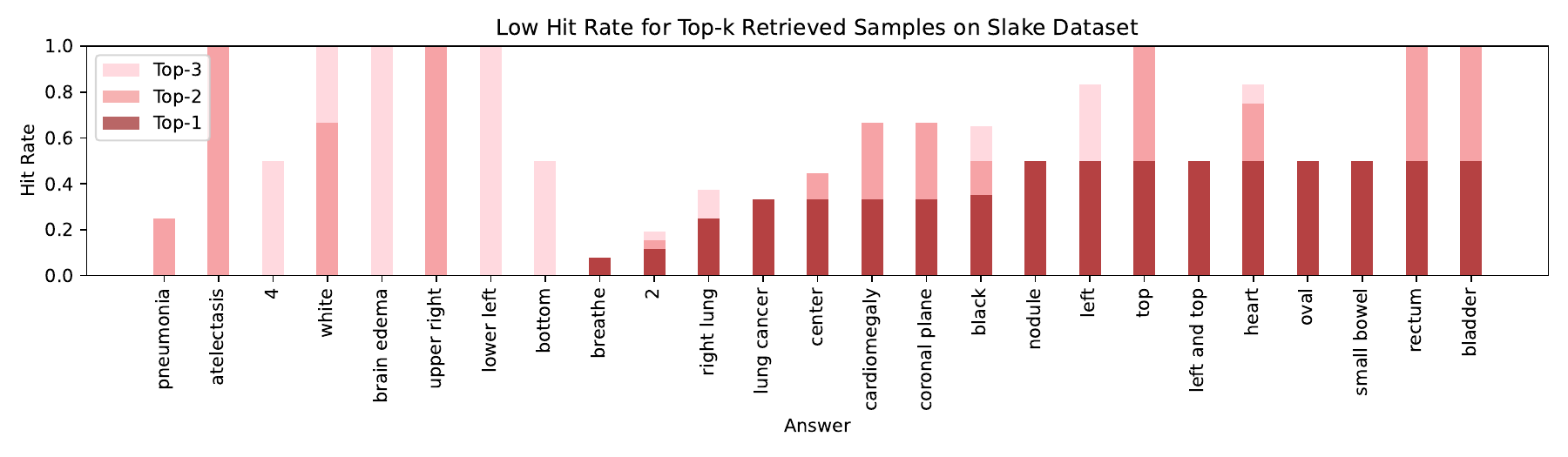}
        \subcaption{}
    \end{minipage}
    \caption{\textbf{Slake dataset}: (a) Entropy Distribution and (b) Overlapping Answer Hit Rate for Top-k.}
    \label{fig:slake_distribution_hit_rate}
\end{figure}

The relationship between entropy and MedVLM outputs was explored by analyzing the high-entropy answers across three MedVQA datasets, as illustrated in Figure \ref{fig:path_distribution_hit_rate}-\ref{fig:slake_distribution_hit_rate} (a). It can be observed that questions involving pronouns like ``this'' and ``their,'' attributes such as ``small'' and ``large,'' numerical or size-related types, and positional references like ``left'' and ``right'' exhibit higher uncertainty. Specifically, for the PathVQA dataset, answers like ``foot,'' ``face,'' ``breast,'' and ``blood'' often correspond to ambiguous questions such as ``What is present?'' Similarly, for the VQA-RAD dataset, inconsistencies between abbreviations like ``pa'' and their full forms ``posterior-anterior'' also contribute to high uncertainty.  \\To further investigate whether entropy correlates with the information present in the dataset, we analyzed the low-hit-rate answers, as illustrated in Figure \ref{fig:path_distribution_hit_rate}-\ref{fig:slake_distribution_hit_rate} (b). A notable correlation between high-entropy answers and low hit rates is observed. For numerical and attribute-related answers, most are not directly retrieved. Moreover, expressions not appearing in the training data, such as ``PA,'' also result in lower hit rates. 

\section{Knowledge Retrieval}

\begin{figure}[h]
    \centering
    \resizebox{1\columnwidth}{!}{%
        \includegraphics{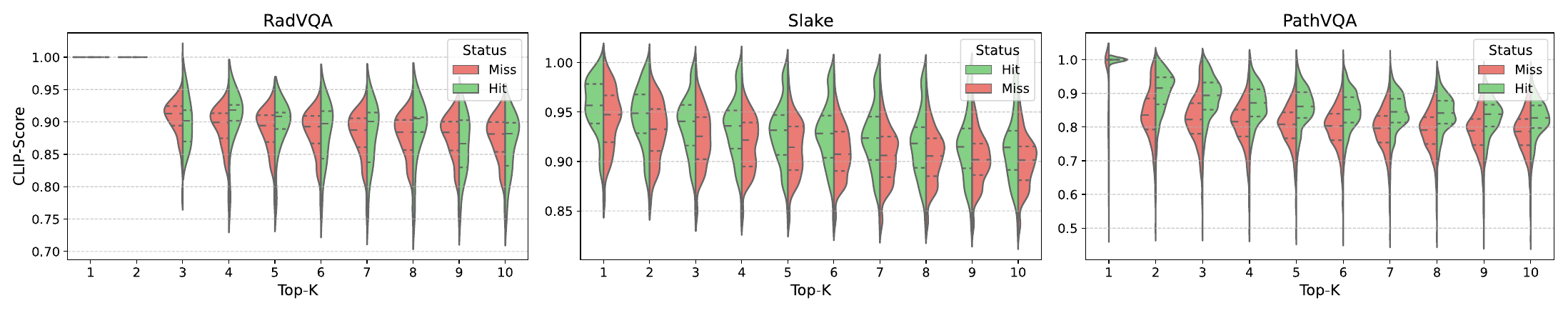}
    }
    \vspace{-0.8cm}
    \caption{\footnotesize CLIP score distributions for open-ended QA hits and misses. }
    \label{fig:clip_score}
\end{figure}

Our knowledge retrieval process utilizes CLIP scores for a preliminary relevance assessment. We rank reference texts by their CLIP scores and select the Top-K most relevant ones. This Top-K strategy is employed instead of applying a fixed score threshold, as any such threshold is highly sensitive to sample variations and lacks generalizability. The selected texts are then merged and provided as input to the GPT-4 API, whose sole function is to extract relevant keywords \textbf{k} for highlighting purposes. Figure \ref{fig:clip_score} illustrates the significant overlap in CLIP scores between documents that contain an explicit answer (``Hits'') and those that do not (``Misses''), demonstrating the unreliability of a simple threshold. It is noteworthy that a 'Miss' does not necessarily imply irrelevance. Our empirical findings on the current datasets suggest that documents with CLIP scores below 0.6 are generally not informative.

\clearpage

\section{Hyperparameter Ablation of Classifier-Free Guidance}
The hyperparameters \(\alpha\), \(\beta\), and \(\gamma\) represent the weights applied to different components in the Expert-CFG framework. Specifically, \(\alpha\) adjusts the weight of the token embeddings in the normal branch, \(\beta\) modulates the attention probability weights in the normal branch, and \(\gamma\) scales the weight applied to the logits. The hyperparameter settings aim to balance the strength of alignment between the original textual output and expert annotations. Table \ref{tab:vqa-hyper-comparison} compares the overall accuracy on the VQA-RAD, SLAKE, and PathVQA datasets for various hyperparameter configurations, highlighting the impact of each parameter on performance. When \(\alpha = 0\), Eq (7) degenerates to fully masking the conditional input, leading to incomplete context and performance degradation.

\begin{table}[htbp]
  \centering
  \resizebox{\textwidth}{!}{
    \begin{minipage}{0.45\textwidth}
      \centering
      \begin{tabular}{l|rrr}
        \toprule
        \multicolumn{4}{c}{\textbf{VQA-RAD Overall Accuracy}} \\
        \midrule
        Phi3.5-V 4.2B & \multicolumn{3}{c}{82.04} \\
        \quad w/ 100\% RAG & \multicolumn{3}{c}{81.15} \\
        \quad w/ 100\% Expert-RAG & \multicolumn{3}{c}{95.34} \\
        \midrule
        \quad w/ 100\% Expert-CFG & \multicolumn{3}{c}{\textbf{Hyper Parameters}} \\
        \midrule
        \rowcolor{gray!20} $\alpha$ (in Eq. (7)) & 0     & 0.01  & 0.1 \\
        \quad($\alpha$, 3.0, 1.3) & 96.23 & \textbf{98.44} & 97.11 \\
        \midrule
        \rowcolor{gray!20} $\beta$ (in Eq. (9)) & 1     & 3     & 5 \\
        \quad(0.01, $\beta$, 1.3) & 98.22 & \textbf{98.44} & 98.00 \\
        \midrule
        \rowcolor{gray!20} $\gamma$ (in Eq. (8)) & 1     & 1.3   & 1.5 \\
        \quad(0.01, 3.0, $\gamma$) & 98.00 & \textbf{98.44} & 97.11 \\
        \bottomrule
      \end{tabular}
      \label{tab:rad-hyper}
    \end{minipage}
    \begin{minipage}{0.45\textwidth}
      \centering
      \begin{tabular}{l|rrr}
        \toprule
        \multicolumn{4}{c}{\textbf{Slake Overall Accuracy}} \\
        \midrule
        Phi3.5-V 4.2B & \multicolumn{3}{c}{85.20} \\
        \quad w/ 100\% RAG & \multicolumn{3}{c}{85.58} \\
        \quad w/ 100\% Expert-RAG & \multicolumn{3}{c}{97.36} \\
        \midrule
        \quad w/ 100\% Expert-CFG & \multicolumn{3}{c}{\textbf{Hyper Parameters}} \\
        \midrule
        \rowcolor{gray!20} $\alpha$ (in Eq. (7)) & 0     & 0.01  & 0.1 \\
        \quad($\alpha$, 3.0, 1.3) & 98.20 & \textbf{99.62} & 98.58 \\
        \midrule
        \rowcolor{gray!20} $\beta$ (in Eq. (9)) & 1     & 3     & 5 \\
        \quad(0.01, $\beta$, 1.3) & \textbf{99.62} & \textbf{99.62} & 99.43 \\
        \midrule
        \rowcolor{gray!20} $\gamma$ (in Eq. (8)) & 1     & 1.3   & 1.5 \\
        \quad(0.01, 3.0, $\gamma$) & 99.24 & \textbf{99.62} & 98.86 \\
        \bottomrule
      \end{tabular}
      \label{tab:slake-hyper}
    \end{minipage}
    \begin{minipage}{0.45\textwidth}
      \centering
      \begin{tabular}{l|rrr}
        \toprule
        \multicolumn{4}{c}{\textbf{PathVQA Overall Accuracy}} \\
        \midrule
        Phi3.5-V 4.2B & \multicolumn{3}{c}{65.24} \\
        \quad w/ 100\% RAG & \multicolumn{3}{c}{72.90} \\
        \quad w/ 100\% Expert-RAG & \multicolumn{3}{c}{88.46} \\
        \midrule
        \quad w/ 100\% Expert-CFG & \multicolumn{3}{c}{\textbf{Hyper Parameters}} \\
        \midrule
        \rowcolor{gray!20} $\alpha$ (in Eq. (7)) & 0     & 0.01  & 0.1 \\
        \quad($\alpha$, 3.0, 1.3) & 91.36 & \textbf{94.76} & 89.94 \\
        \midrule
        \rowcolor{gray!20} $\beta$ (in Eq. (9)) & 1     & 3     & 5 \\
        \quad(0.01, $\beta$, 1.3) & 93.64 & \textbf{94.76} & 92.63 \\
        \midrule
        \rowcolor{gray!20} $\gamma$ (in Eq. (8)) & 1     & 1.3   & 1.5 \\
        \quad(0.01, 3.0, $\gamma$) & 93.09 & \textbf{94.76} & 92.03 \\
        \bottomrule
      \end{tabular}
      \label{tab:path-hyper}
    \end{minipage}
  }
  \caption{Comparison of Overall Accuracy with Different Hyperparameter Settings for VQA-RAD, Slake, and PathVQA.}
  \label{tab:vqa-hyper-comparison}
\end{table}

\section{Zero-Shot Cases of Baseline MedVLM and Expert-CFG}
We conducted additional zero-shot case studies on images and questions from three datasets: VQA-RAD, SLAKE, and PathVQA. While the baseline MedVLM's responses are relevant to the input images, they exhibit certain errors due to a lack of knowledge or fine-grained understanding. For instance, the baseline model fails to distinguish between ``fat'' and ``fat stranding," recognize fine-grained details such as EKG electrodes, or differentiate between ``T1-weighted MRI" and ``T2-weighted MRI,'' among other issues. These limitations highlight the importance of incorporating external knowledge and expert guidance for accurate and clinically reliable outputs.

\begin{figure*}[htbp]
    \centering
    \includegraphics[width=\textwidth]{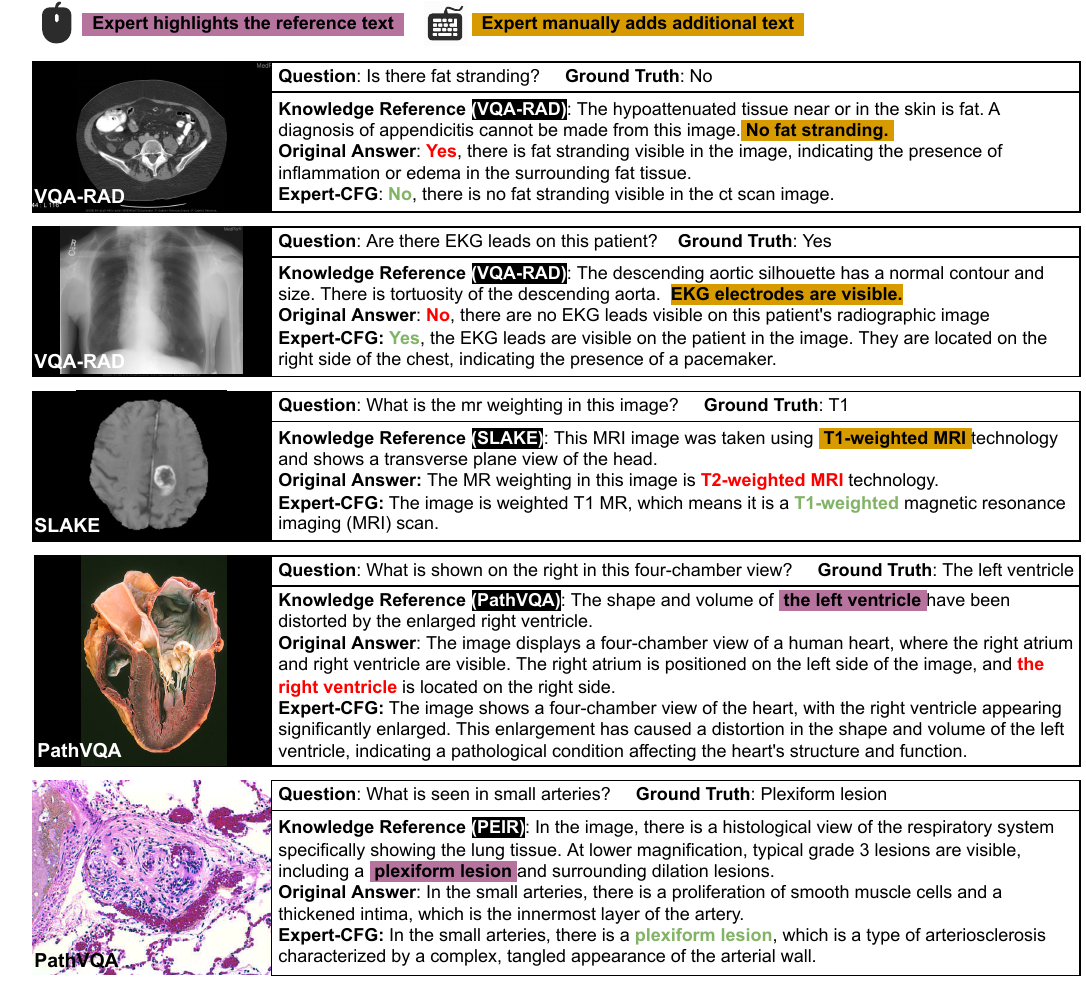}
    \caption{\textbf{Zero-shot case study examples from three datasets: VQA-RAD, SLAKE, and PathVQA.}}
    \label{fig:case_study}
\end{figure*}

\clearpage

\section{Additional Zero-Shot Chinese Capability of Baseline MedVLM and Expert-CFG}
\begin{figure*}[htbp]
    \centering
    \includegraphics[width=\textwidth]{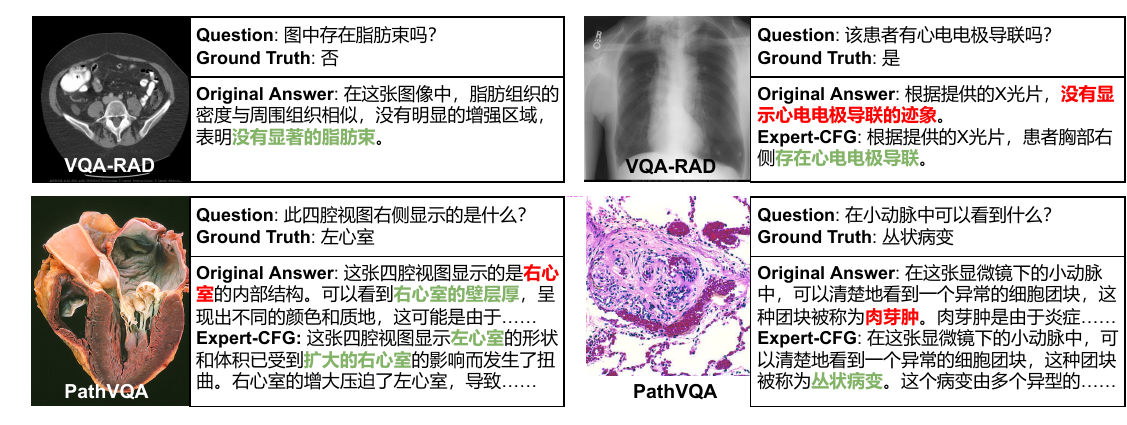}
    \caption{\textbf{Additional zero-shot chinese case study examples from three datasets: VQA-RAD, SLAKE, and PathVQA.}}
    \label{fig:case_study_chi}
\end{figure*}

We were surprised to discover that Phi3V-Med demonstrated zero-shot capability in Chinese as shown in Figure \ref{fig:case_study_chi}, although the baseline outputs in Chinese and English for the same question were inconsistent. For example, the Chinese response was able to correctly recognize ``fat stranding," whereas the English response failed to do so. Expert-CFG also proved effective in incorporating expert annotations in Chinese.

\clearpage
\section{Low-Entropy Error Analysis and Mitigation Pathways}

While our uncertainty estimation achieves AUC $>$ 0.8 across benchmarks, persistent low-entropy errors reveal critical directions for improvement. Excluding ambiguous cases such as questions like ``What does this image show?" and answers like ``CT" or its full name, errors predominantly stem from anatomical positional confusion. For example, the inability to distinguish left from right due to the distinction between PA/AP views accounts for 21.73\% and 42.02\% in the VQA-RAD and SLAKE datasets respectively. In addition, there are several other issues. There is a lack of the concept of degree (such as ``larger", ``mild"), and there is an inability to accurately handle numbers, which is an inherent problem of VLM. Disease misinterpretation accounts for 17.39\% and 14.49\%, and unlearned medical concepts (such as words that do not appear in the training set) account for 26.08\% and 7.24\% in the relevant datasets respectively. We attribute these to sparse visual-textual grounding in training data. For instance, insufficient annotations for radiographic projection labels hinder left-right discrimination. Addressing such issues requires fine-grained synthetic data augmentation (such as view-specific anatomical templates) and knowledge-anchored multi-turn dialogue to reinforce spatial reasoning.

\begin{table}[htbp]
  \centering
  \resizebox{\linewidth}{!}{
    \begin{tabular}{llllcc}
    \toprule
    \textbf{Type}  & \multicolumn{3}{c}{\textbf{Examples of low entropy ($<$0.35) but incorrect results}} & \multicolumn{2}{c}{\textbf{Proportion (\%)}} \\
\cmidrule{2-6}          & Question & GT & Prediction (Entropy) & VQA-RAD (23) & SLAKE (69) \\
    \midrule
    \makecell[l]{Question\\ ambiguity} & What is under the right hemidiaphragm? & free air & \makecell[l]{stomach \\ bubble (0.011)} & 4.34 & 7.24 \\
    \midrule
    \makecell[l]{Synonymous\\ answers} & What modality is used to take this image? & xr    & x-ray (0.013) & 26.08 & 2.89 \\
    \midrule
    Position & Which side is more clearly visualized? & left  & right (0.032) & 21.73 & 42.02 \\
    \midrule
    Degree & \makecell[l]{Is the heart size in this image smaller \\ or larger than if the image was taken AP?} & smaller & larger (0.031) & 4.34 & 5.79 \\
    \midrule
    Disease & What is abnormal about the pancreas? & enlarged & \makecell[l]{fatty \\ infiltration (0.078)} & 17.39 & 14.49 \\
    \midrule
    Numbers & How many lungs have existed in this image? & 2     & 1 (0.2524) & 0.00 & 18.84 \\
    \midrule
    Shapes & What is the shape of larynx in this image? & oval  & irregular(0.0046) & 0.00 & 1.44 \\
    \midrule
    \makecell[l]{Not appeared in \\ the training set} & What organ system is the pathology? & lymphatic & cardiovascular (0.329) & 26.08 & 7.24 \\
    \bottomrule
    \end{tabular}%
  }
  \caption{\textbf{Examples of Low-entropy ($<$0.35) incorrect results.}}
  \label{tab:addlabel}%
\end{table}%

\section{Limitations and Future Works}
Despite the fact that our framework has achieved state-of-the-art performance on three MedVQA datasets, due to the significant gaps in different sub-fields within the medical domain, it is necessary to further expand the experimental evaluation. Besides, the greedy decoding strategy may not be globally optimal for long sequences. However, currently, there is a lack of long-sequence benchmarks in the medical field for further research on the estimation of uncertainty in and knowledge representation of MedVLMs. Meanwhile, the framework we proposed provides the medical community with a cost-effective method to align MedVLMs with expert knowledge. It can be combined with high-accuracy classifiers or detectors to synthesize data to further improve the performance and generalization ability of the model.

\end{document}